# DeepSTCL: A Deep Spatio-temporal ConvLSTM for Travel Demand Prediction

Dongjie Wang
*School of Information Science and Technology*
*Southwest Jiaotong University*
Chengdu, China
wangdongjiescu@foxmail.com

Yan Yang*, *Member, IEEE*
*School of Information Science and Technology*
*Southwest Jiaotong University*
Chengdu, China
yyang@swjtu.edu.cn

Shangming Ning
*School of Information Science and Technology*
*Southwest Jiaotong University*
Chengdu, China
ningsm624@gmail.com

*Abstract*—Urban resource scheduling is an important part of the development of a smart city, and transportation resources are the main components of urban resources. Currently, a series of problems with transportation resources such as unbalanced distribution and road congestion disrupt the scheduling discipline. Therefore, it is significant to predict travel demand for urban resource dispatching. Previously, the traditional time series models were used to forecast travel demand, such as AR, ARIMA and so on. However, the prediction efficiency of these methods is poor and the training time is too long. In order to improve the performance, deep learning is used to assist prediction. But most of the deep learning methods only utilize temporal dependence or spatial dependence of data in the forecasting process. To address these limitations, a novel deep learning traffic demand forecasting framework which based on Deep Spatio-Temporal ConvLSTM is proposed in this paper. In order to evaluate the performance of the framework, an end-to-end deep learning system is designed and a real dataset is used. Furthermore, the proposed method can capture temporal dependence and spatial dependence simultaneously. The closeness, period and trend components of spatio-temporal data are used in three predicted branches. These branches have the same network structures, but do not share weights. Then a linear fusion method is used to get the final result. Finally, the experimental results on DIDI order dataset of Chengdu demonstrate that our method outperforms traditional models with accuracy and speed.

## I. INTRODUCTION AND RELATED WORK

Travel demand prediction is an important task of traffic resources scheduling. It is common for highway traffic resources schedule problems in real life, which mainly include public transportation resources and taxi transportation resources. Generally, the public transportation resources are always chosen by people because of convenience and low price. However, with the development of society, the public transportation resources present their disadvantages such as highly time costing and hardly meeting people's needs [1]. Hence, many new transportation modes appear, such as DIDI, UBER and etc. It is easy for passengers to use the mobile application to book a taxi. In this situation, a lot of order data will be generated. These data are able to reflect the trend of travel demand in the short term. Thus, the order data is used to predict travel demand, achieve appropriate urban resource scheduling and provide better services for passengers in this mode. In this paper, a Deep Spatio-Temporal Convolutional LSTM (DeepSTCL) is proposed to forecast travel demand which considers the time and space factors comprehensively and gets a great prediction performance.

Travel demand data is typical spatio-temporal data. This type of data has obviously periodicity, so historical travel demand is used to predict future travel demand. However, the traditional time series model has many deficiencies. For example, if the prediction span is large, the model performance will be worse [2]; if the data missing problem is serious, the model still cannot get a good result. Travel demand data also have these characteristics. Therefore, how to balance data and make full use of spatio-temporal factors is very important in the prediction process.

There is a large number of classic methods in traditional time-series data mining, such as AR, MA, and ARMA etc. The AR model is relatively simple which considers the future time points can be predicted by the linear combination of historical time points. The MA model is similar to AR model, but MA is the linear combination of white noise; The ARMA model contains the characteristics of AR and MA [3]. In addition, some improved models like ARIMA and SARIMA are proposed. These models are fit for different scenarios. However, if the application scenarios are complex and the size of data is large, these methods will become very difficult to get a good result and their performance will be worse. Nowadays, more and more machine learning and deep learning models are introduced into the spatio-temporal learning field, such as

*Corresponding author: yyang@swjtu.edu.cn. This work is supported by the National Natural Science Foundation of China (No. 61572407), the Soft Science Foundation of Sichuan Province (No. 2016ZR0034), the National Key Research and Development Program of China (No. 2016YFC0802209) and the National Natural Science Foundation of China (No. 61773324).

978-1-5090-6014-6/18/$31.00 ©2018 IEEE

neural networks, perceptron and SVM [4]. These models have better performance and higher accuracy than traditional models.

In the past, many researchers in the field of transportation only used the temporal dependence of spatio-temporal data [5], but ignored spatial dependence. Therefore, the prediction ability of these methods could be improved greatly. Li et al. [6] used a modified ARIMA model to predict travel demand and the experimental performance is good. A multi-view clustering algorithm was discussed by Davis et al. [7] to aggregate travel demand from multiple perspectives and exploit the unsupervised model to predict travel demand. Chidlovskii et al. [8] employed a multi-task learning framework to predict travel demand and achieved good results, but the data information was not fully tapped. When the amount of data becomes larger, the model will become complicated.

Recently, deep learning has made remarkable achievements in various fields. Sutskever et al [9] proposed a method that improved the accuracy of the translation from English to French significantly, which using multi-layers LSTM to encode and decode word vectors. Kalchbrenner et al [10] studied a dynamic convolutional neural network (DCNN) which improved Facebook's emotional prediction accuracy to 75%. Shi et al [11] presented a Convolutional LSTM neural network that had better prediction performance on handwriting digital recognition; Kim et al [12] utilized ConvLSTM to predict rainfall that was superior to traditional methods. Therefore, deep learning is suitable to assist prediction.

In order to provide smart decisions to traffic resource scheduling, the novel deep learning techniques become a key tool of travel demand prediction. For example, Lv et al [13] utilized a deep auto-encoder to predict traffic flow. The feasibility of the deep model in traffic prediction was proved. However, due to the limitations of the unsupervised learning model, the accuracy of this method was not good; Ma et al [14] regarded the traffic flow as an image and used CNN to solve prediction, but this model lacked temporal dimension information, so it might be improved greatly. A deep stack auto-encoder model was investigated by Duan et al [15] to fill the traffic missing value. The same as Lv, this accuracy of the unsupervised model was disappointed. So we want to put forward an end-to-end learning method that predicts travel demand accurately and efficiently.

Our DeepSTCL is an end-to-end travel demand forecasting framework. It regards the historical travel data as a video stream, and then uses the ConvLSTM network to predict travel demand. Due to the ConvLSTM network simultaneously exploits the spatio-temporal feature of the data, our network can be applied to the field of spatio-temporal data. In addition, the experimental results have shown that the framework has a low time cost.

In this paper, our contribution is summarized as follows:

1. An end-to-end travel demand forecasting framework (DeepSTCL) is proposed. This method makes full use of the spatio-temporal factors of the data and achieves great performance in accuracy and time.

2. Visualization technology is used to compare real results and forecast results. It gets a clear comparison of prediction effect.

3. The effectiveness of different size partitioning schemes on accuracy and time are analyzed.

The remainder of the paper is structured as follows. The problem definition and preliminaries are adopted in Section II. In Section III, the DeepSTCL framework and the other two baseline neural network frameworks (LSTM, CNN) are depicted. And the similarities and differences between the three frameworks are compared. Section IV presents the criteria of evaluation and the results of experiments. It reflects the effectiveness and accuracy of DeepSTCL. Section V concludes this paper.

## II. PRELIMINARIES

In this section, the detailed definitions of travel demand are introduced in subsection II-A and that of demand prediction problems is formally depicted in subsection II-B.

### A. Detailed definitions

Since the overall travel demand of the entire city is a summary of every part, so we divide the entire city into many blocks to predict. Vehicle scheduling decision will be smartly made according to travel demand.

**Definition 1 (Geographical rectangle).** In this work, we divide the latitude and longitude into $n \times n$ geographical rectangles [16]. The geographical area is divided in Fig. 1.

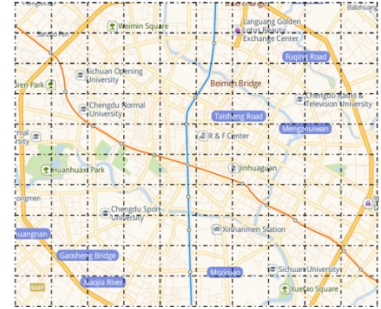

Fig. 1. Geographical rectangles

By this way, the overall travel demand about one city is reflected by each block's demand. In addition, the travel demand of block is not only related to local heat, but also associated with neighboring blocks.

**Definition 2 (Travel demand).** Travel demand [17] is defined as follows:

$$TD_k^{m,n} = \sum_{OD_k \in R} |\{R(m,n)\}| \qquad (1)$$

Where, $R(m, n)$ means the geographical rectangle $(m, n)$; $|.|$ is the cardinal of $a$ set; $OD_k \in R$ represents the $k^{th}$ timestamp belonging to R; $TD_k^{m,n}$ means the order count at $k^{th}$ timestamp in R, as shown in Fig. 2.

| 3 | 4 | ... | 6 | 5 |
| --- | --- | --- | --- | --- |
| 6 | 99 | ... | 98 | 6 |
| ... | ... | ... | ... | ... |
| 7 | 96 | ... | 99 | 4 |
| 4 | 89 | ... | 84 | 1 |

Fig. 2. Order Count at $k^{th}$ time

The definition of travel demand has been given. The travel demand at a moment is regarded as a snapshot. This snapshot is an M × N matrix. Snapshots over a period of time form a snapshot stream. It can be represented by a tensor $X \in \mathbb{R}^{t \times M \times N}$.

### B. Problem Definition

In this paper, we try to predict the travel demand based on the historical travel demand of the whole area for every moment in different regions.

**Problem.** Given the historical observations $TD_k^{m,n}$ for $k$ $t-p, \ldots, t-1$, predict $TD_t^{m,n}$.

## III. MODELS

In this section, the DeepSTCL framework is introduced in subsection III-A. Then Convolutional Neural Network is shown in subsection III-B, and LSTM Network is presented in subsection III-C.

### A. Deep Spatial-Temporal Convolutional LSTM Network

Fig. 3 illustrates the architecture of Deep Spatial-Temporal Convolutional LSTM Network (DeepSTCL), which consists of three components: closeness, period and trend. For closeness, as Fig. 3 shows, in the first step, the latitude and longitude are divided into $n$ parts respectively. Then the order count in different zones is added up and it is filled into geographical rectangles. Eventually the historical observation for the entire region is obtained such as $TD_{t-k}, TD_{t-k-1}, \ldots, TD_{t-1}$. In this paper, the travel demand over a period time is regarded as a video stream. Therefore, video prediction method can be used to predict short-term travel demand in the future. Firstly, a travel demand video stream is fed into the prediction framework; Secondly, the predicted intermediate results are fed into a relu-activated three-dimensional convolutional layer; Thirdly, the predicted video segment $TD_{t-k-1}, TD_{t-k-2}, \ldots, TD_t$ of a single branch is obtained. The last snapshot of this video segment is the prediction of this branch. The same process is applicable to the period branch and trend branch. Ultimately, the final prediction is got by linear fusion.

In the field of traditional spatio-temporal, a long historical observation is an input into the model. It is difficult for a traditional time-series method to obtain the spatio-temporal dependency in the dataset. In addition, the temporal attributes of different spans have a great influence on the result of the prediction. So through the knowledge of the spatiotemporal domain, the higher-dependent temporal feature is extracted from three aspects (closeness, period, trend) [18]. There are two benefits in this method. One is that reduces the volume of data and decreases data noise. The other is that it improves the robustness and accuracy of the model.

In this model, ConvLSTM is used to capture spatial-temporal dependency in the dataset. The differences between ConvLSTM and LSTM are that ConvLSTM changes the feedforward method of LSTM from Hadamard product to convolution. The main formula for ConvLSTM is shown as follows:

$$i_t = \sigma(W_{xi} * \mathcal{X}_t + W_{hi} * \mathcal{H}_{t-1} + W_{ci} \circ \mathcal{C}_{t-1} + b_i) \quad (2)$$

$$f_t = \sigma(W_{xf} * \mathcal{X}_t + W_{hf} * \mathcal{K}_{t-1} + W_{cf} \circ \mathcal{C}_{t-1} + b_f) \quad (3)$$

$$\mathcal{C}_t = f_t \circ \mathcal{C}_{t-1} + i_t \circ tanh(W_{xc} * \mathcal{X}_t + W_{hc} * \mathcal{H}_{t-1} + b_c) \quad (4)$$

$$o_t = \sigma(W_{xo} * \mathcal{X}_t + W_{ho} * \mathcal{K}_{t-1} + W_{co} \circ \mathcal{C}_t + b_o) \quad (5)$$

$$\mathcal{H}_t = o_t \circ \tanh(\mathcal{C}_t) \quad (6)$$

where, '*' represents the convolution operation and 'o' represents the Hadamard product. As above formulas shown, LSTM is the special form of ConvLSTM, so the ConvLSTM can be used like LSTM. At the same time, Zhang et al [19] found that one convolutional layer depicts the concept of distance. Two convolutional layers describe the relationship between adjacent areas. Hence, more ConvLSTM are suitable to handle large dataset. Because the convolutional property of these ConvLSTM layers are able to capture more spatial-temporal features.

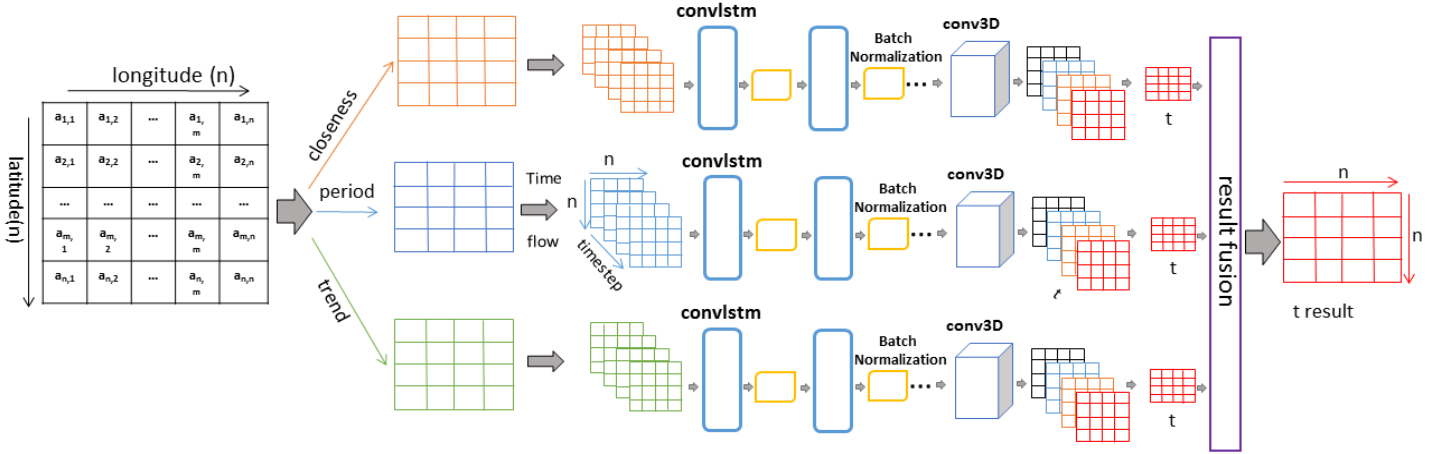

Fig. 3. Deep Spatial-Temporal Convolutional LSTM Network

Generally speaking, ConvLSTM consists of two network structures for forecasting issues, one is the encoding network and the other is the forecasting network [11]. As is shown in Fig. 4, the encoding layer and the prediction layer share the middle parts. And the predicted result and the actual result have the same form. In this paper, $[TD_{t-p}, TD_{t-p-1}, ..., TD_{t-1}]$ is fed into DeepSTCL, $[TD_{t-p-1}, TD_{t-p-2}, ..., TD_t]$ is the final output. The sequence $[TD_{t-p-1}, TD_{t-p-2}, ..., TD_{t-1}]$ plays the role of middle layers between the encoding network and the forecasting network. So the process of encoding and decoding share the same middle network structure.

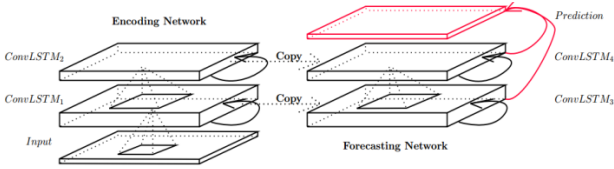

Fig. 4. The structure of encoding-forecasting of ConvLSTM network [11]

Here, a Batch Normalization layer is used in DeepSTCL. It accelerates the speed of training and avoids over-fitting [20]. The formulas for this process are shown as follows:

$$\hat{x}^{(k)} = \frac{x^{(k)} - E[x^{(k)}]}{\sqrt{Var[x^{(k)}]}} \quad (7)$$

$$y^{(k)} = \gamma^{(k)} \hat{x}^{(k)} + \beta^k \quad (8)$$

In our framework, the Batch Normalization layer is behind ConvLSTM layer. It normalizes the output of ConvLSTM. But Eq. (7) will reduce the expression ability of the model, so it uses Eq. (8) to keep the nonlinear properties of the model. Naturally, this layer makes the model more quick and reliable.

This framework has three components (i.e. closeness, period, and trend) [21]. The linear weighting method is used to fuse three results in the final step. The fusion equation is shown as follows:

$$TD_t = W_c \circ TD_{t_c} + W_p \circ TD_{t_p} + W_t \circ TD_{t_t} \quad (9)$$

where $W_c$, $W_p$, $W_t$ are the weight matrix of closeness, period and trend respectively; "∘" represents the Hadamard product; which multiplies two matrices by element-wise. $TD_{t_c}$ is the closeness model's output; $TD_{t_p}$ is the period model's output; $TD_{t_t}$ is the trend model's output; $TD_t$ is the fusion result.

The whole training process of DeepSTCL is shown in Algorithm1. We firstly construct a different historical snapshot sequence as a video stream. Then, DeepSTCL is trained by backpropagation and Adam.

**Algorithm 1** DeepSTCL Training Algorithm

**Require:** Historical observations: $\{X_0, \cdots, X_{n-1}\}$;
Historical snapshot sequence: $\{TD_0, \cdots, TD_{n-1}\}$;
Snapshot width: $n_{lng}$, Snapshot height: $n_{lat}$;
Sequence length: $s_l$;
Lengths of closeness, period, trend branches: $l_c, l_p, l_q$;
closeness: $c$, period: $p$, trend: $t$;

**Ensure:** DeepSTCL Model
1: **for** all available time interval $t(1 \leq t \leq n-1)$ **do**
2:   $S_c = [TD_{t-l_c}, TD_{t-(l_c-1)}, \cdots, TD_{t-1}]$
3:   $S_p = [TD_{t-l_p}, TD_{t-(l_p-1)}, \cdots, TD_{t-p}]$
4:   $S_q = [TD_{t-l_q}, TD_{t-(l_q-1)}, \cdots, TD_{t-q}]$
    // $TD_t$ is the target at time t
    put $S_c, S_p, S_q$ into DeepSTCL $B_c, B_p, B_q$ respectively.
5: **end for**
6: **repeat**
7:   randomly select different batch and dropout find best $\theta$ to make the loss least.
8: **until** stopping criteria is met

### B. Convolutional Neural Network Travel Demand

In subsection III-A, DeepSTCL has been proposed for travel demand prediction. Now one baseline model, named Convolutional Neural Network Travel Demand is introduced in subsection III-B. In this framework, spatial dependency is only used to predict the travel demand.

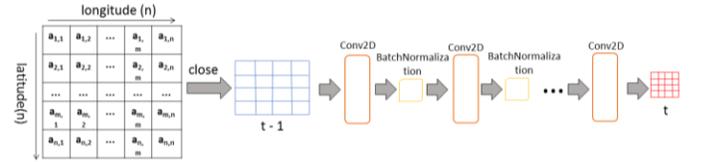

Fig. 5. Convolutional Neural Network Travel Demand

Fig. 5 shows the framework of a convolutional neural network. Similar to DeepSTCL, firstly the longitude and latitude are divided into *n* parts; secondly the snapshot at *t-1* time point is regarded as the input of the model; finally the prediction snapshot of model is the final result. This predicted structure only uses spatial dependency in the dataset. It does not consider the temporal dependency of the dataset.

### C. Long Short-term Neural Network Travel Demand

Long Short-term Neural Network Travel Demand is another baseline framework, which is brought up in this subsection. Long Short-term Neural Network is the main part of this framework. This neural network is a complicated recurrent neural network. There are four transmission gates in one LSTM module. These gates are used to solve the disappearance of the gradient in the long context [22]. Fig. 6 shows an instance of LSTM module.

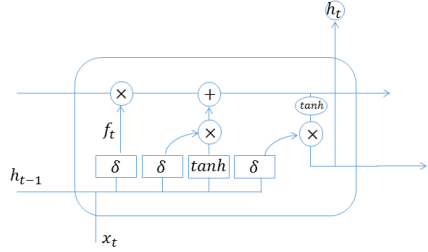

Fig.6. an instance of LSTM module

An LSTM network is used to predict the travel demand of a small area. The prediction process of LSTM is shown in Fig. 7.

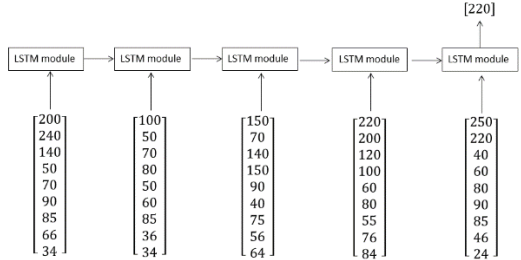

Fig. 7. Structure of LSTM Travel Demand

As Fig. 7 shows, the travel demand for an area at the $t-n$ moment is delivered to the first LSTM module. Then, the next step travel demand for this area is given to the second LSTM module. This operation is repeated to the last LSTM module. The output of the last LSTM module is the forecast value in one area. In the end, the final result is gotten by filling the prediction of all regions according to geographic location. This structure characterizes the temporal dependency, but it doesn't utilize the spatial dependency between different blocks.

## IV. EXPERIMENTAL EVALUATION

In order to compare the pros and cons of our framework with the other baseline frameworks, the experiments are conducted on a real dataset. In subsection IV-A, the network structures of our framework and baseline frameworks are shown. In subsection IV-B, dataset and data preprocessing process are introduced. In subsection IV-C, the experimental results, evaluation indicators, and analysis of results are elaborated.

### A. Models

According to different temporal dependencies, our model consists of four different situations, including Convolutional LSTM Closeness (CLC), Convolutional LSTM Period (CLP), Convolutional LSTM Trend (CLT) and Deep Spatiotemporal Convolutional LSTM (DeepSTCL). LSTM and CNN are baseline models. LSTM only considers temporal dependency while CNN focuses on spatial dependency. DeepSTCL is the fusion of CLC, CLP, and CLT.

In this experiment, LSTM and CNN have different network structures. The four cases of our model share the same network structure. Their structures are shown in Table I.

TABLE I.  DIFFERENT MODEL STRUCTURES IN EXPERIMENT

| Model | Layer Type | Input | Output | Filters | Kernel Size | Drop out |
|---|---|---|---|---|---|---|
| LSTM | LSTM1 | 10*10 | 100*10 | ---- | ---- | 0.2 |
|  | LSTM2 | 100*10 | 1 | ---- | ---- | 0.2 |
| CNN | Conv2D1 | 74*74 | 74*74 | 74 | 3*3 | ---- |
|  | BatchNormalization | ---- | ---- | ---- | ---- | ---- |
|  | Conv2D2 | 74*74 | 74*74 | 74 | 3*3 | ---- |
|  | BatchNormalization | ---- | ---- | ---- | ---- | ---- |
|  | Conv2D3 | 74*74 | 74 | 1 | 3*3 | ---- |
| Deep STCL | ConvLSTM2D1 | 74*74 | 74*74 | 74 | 3*3 | 0.13 |
|  | BatchNormalization | ---- | ---- | ---- | ---- | ---- |
|  | ConvLSTM2D2 | 74*74 | 74*74 | 74 | 3*3 | 0.13 |
|  | BatchNormalization | ---- | ---- | ---- | ---- | ---- |
|  | Conv3d | 74*74 | 74 | 1 | 3*3*5 | ---- |

### B. Dataset and preprocessing

The dataset is DIDI order data which supplied by DIDI Gaia Plan [23]. The dataset covers all order data and GPS data in Chengdu in November 2016. Each order in the data contains order ID, start billing time, end billing time, pick-up, drop-off longitude, and latitude.

Firstly, latitude and longitude are divided into 74*74 parts, 5476 geographical blocks can be gotten. The area of each block is 1 km$^2$. Secondly, within a span of one hour, the travel demand in the different region is counted. However, the travel demand in different regions is not continuous, so the missing value in the regions are filled with zero.

When it is required to feed data into LSTM Neural Network, each order number is matched with each area. Different regions have different input sequences in the following form:

$$TD_k^{m,n} = TD_1^{m,n}, TD_2^{m,n}, \ldots, TD_k^{m,n} \qquad (9)$$

The data should be converted into a matrix form when they need to be feed data ConvLSTM or CNN. Therefore, the data format is listed as follows:

$$\begin{bmatrix} 10 & 50 & 150 & \cdots & \cdots & 200 & 20 & 10 \\ 15 & 60 & 130 & \cdots & \cdots & 220 & 20 & 90 \\ 5 & 45 & 110 & \cdots & \cdots & 145 & 20 & 10 \\ \cdots & \cdots & \cdots & \cdots & \cdots & \cdots & \cdots & \cdots \\ 9 & 65 & 105 & \cdots & \cdots & 134 & 20 & 10 \\ 8 & 43 & 150 & \cdots & \cdots & 156 & 20 & 10 \\ 10 & 50 & 160 & \cdots & \cdots & 130 & 20 & 10 \end{bmatrix}$$

This matrix composed of 74 rows and 74 columns and each number indicates area order number.

### C. Experiments result and Evaluation

In this experiment, we divided the dataset into two parts based on time attributes: test set and train set. The test set is the last three days of November and the train set is all the other days in November.

There are three evaluation metrics in this experiment, which are Root Mean Square Error (RMSE), Mean Absolute Error (MAE) and Mean Absolute Percentage Error (MAPE).

$$RMSE = \sqrt{\frac{1}{u} \times \sum_i (TD_i - \widehat{TD}_i)^2} \qquad (10)$$

$$MAE = \frac{1}{u} \times \sum_i |TD_i - \widehat{TD}_i| \qquad (11)$$

$$MAPE = \frac{100}{u} \times \sum_i \left|\frac{TD_i - \widehat{TD}_i}{TD_i}\right| \qquad (12)$$

where $TD_i$ and $\widehat{TD}_i$ are the true value and predict value, $u$ is the number of geographical rectangles.

In order to evaluate our frameworks, we deploy our code on AMAX Calculate Solution. It has Intel (R) Xeon (R) CPU E5-2620 v4 @ 2.10GHz * 32, 64GB RAM and NVIDIA TITAN XP * 4. Our deep learning model realizing by Keras with Tensorflow.

The RMSE, MAE and MAPE in the experiment are shown in Table II, Table III and Table IV respectively.

TABLE II  RMSE OF SIX MODELS.

| time / Models | 6:00 | 9:00 | 12:00 | 15:00 | 17:00 | 20:00 | 23:00 |
|---|---|---|---|---|---|---|---|
| LSTM | 1.74 | 4.23 | 3.58 | 4.01 | 3.89 | 4.26 | 3.09 |
| CNN | 1.43 | 3.65 | 3.22 | 3.61 | 3.44 | 3.74 | 3.74 |
| CLC | 1.22 | 3.18 | 3.15 | 3.64 | 3.56 | 3.84 | 3.63 |
| CLP | 1.20 | 4.61 | 3.22 | 3.36 | 3.64 | 3.26 | 2.74 |
| CLT | 1.28 | 3.37 | 3.07 | 3.42 | 3.33 | 3.08 | 3.13 |
| **DeepSTCL** | **1.20** | **3.16** | **2.92** | **3.22** | **3.15** | **2.97** | **2.72** |

TABLE III  MAE OF SIX MODELS.

| time / Models | 6:00 | 9:00 | 12:00 | 15:00 | 17:00 | 20:00 | 23:00 |
|---|---|---|---|---|---|---|---|
| LSTM | 0.89 | 1.53 | 1.18 | 1.19 | 1.32 | 1.19 | 0.80 |
| CNN | 0.50 | 1.61 | 1.20 | 1.19 | 1.28 | 1.22 | 1.02 |
| CLC | 0.38 | 1.15 | 1.16 | 1.25 | 1.22 | 1.22 | 0.97 |
| CLP | 0.37 | 1.51 | 1.18 | 1.18 | 1.22 | 1.10 | 0.80 |
| CLT | 0.38 | 1.27 | 1.07 | 1.20 | 1.15 | 1.07 | 0.83 |
| **DeepSTCL** | **0.35** | **1.10** | **1.04** | **1.11** | **1.10** | **0.96** | **0.78** |

TABLE IV  MAPE OF SIX MODELS.

| time / Models | 6:00 | 9:00 | 12:00 | 15:00 | 17:00 | 20:00 | 23:00 |
|---|---|---|---|---|---|---|---|
| LSTM | 18.23 | 33.46 | 32.61 | 32.08 | 31.04 | 28.03 | 20.08 |
| CNN | 13.86 | 24.10 | 22.79 | 22.79 | 21.71 | 18.86 | 15.24 |
| CLC | 11.50 | 18.57 | 22.65 | 20.67 | 22.35 | 20.13 | 15.40 |
| CLP | 12.35 | 20.26 | 23.26 | 20.25 | 20.89 | 18.90 | 14.55 |
| CLT | 11.88 | 18.84 | 22.09 | 20.19 | 21.18 | 19.10 | 14.44 |
| **DeepSTCL** | **11.21** | **18.49** | **21.09** | **19.35** | **20.57** | **17.89** | **14.02** |

From Table II, it is noted that the value of RMSE affects by time points. The reason is that different regions have different travel demand at different times. RMSE is smaller at 6:00, because the travel demand is smaller in a whole day. At the same time, it is noted that DeepSTCL framework outperforms than other frameworks at any time points.

What's more, MAE measures the accuracy of the model, and MAPE eliminates the impact of dimension and reflects the stability and accuracy of the model. Therefore, Table III and IV depict that DeepSTCL is superior to other frameworks in terms of stability and accuracy.

In order to predict deeply in one day, we plot a line chart to present prediction result and true travel demand in one day. From Fig. 8, DeepSTCL not only has the highest accuracy but also has a trend similar to real data. LSTM model has excellent performance in the initial period, but the forecast result becomes disappointed and the fluctuations of prediction are relatively heavy after 12:00. The overall forecasting tendency of LSTM is inferior to the DeepSTCL. CNN has high accuracy in the first half of the day while it cannot characterize the overall trend of data, either.

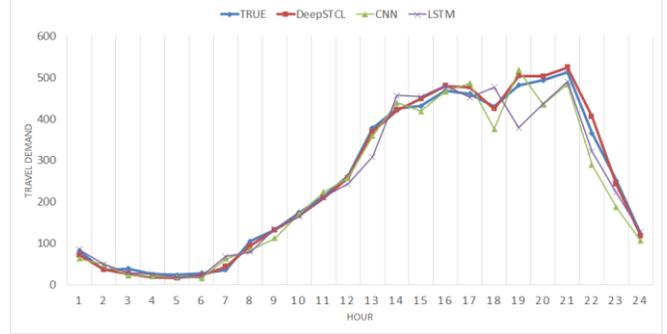

Fig. 8. Travel demand in one day for a single address.

The heat maps at different peak time in real and DeepSTCL predicted situations are shown in Fig. 9. It is worth noting that our predicted result is able to reflect actual demand situation. Here, the maximum values at three peak time are not the same. The max values of 9:00, 12:00, 17:00 are 220, 369.5, 495 respectively. At 9:00, the travel demand for each region is a balance. After the afternoon, the distribution of travel demand mainly spreads from the center which is Chun Xi Road. It is noted from these pictures, the prediction of DeepSTCL is accurate at different time points. It indicates that DeepSTCL has excellent performance under different data imbalance situation.

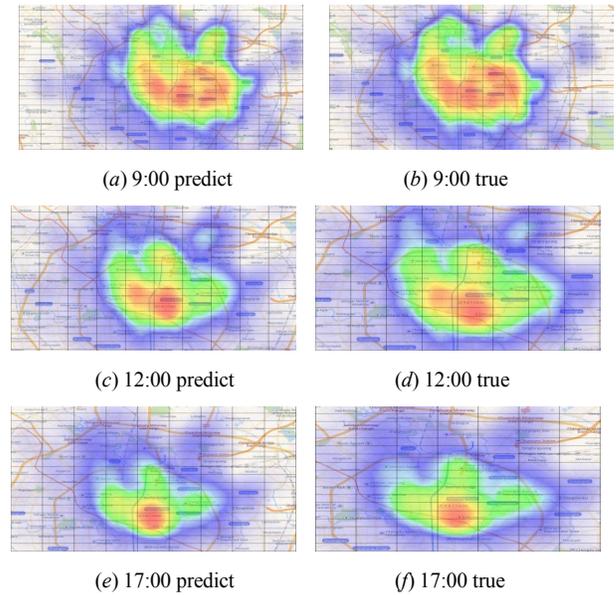

(a) 9:00 predict  (b) 9:00 true

(c) 12:00 predict  (d) 12:00 true

(e) 17:00 predict  (f) 17:00 true

Fig. 9. The heat map at different peak time in a different region

In our data processing, we split the geographical area into n*n blocks. Discovered through experiments, the size of each block affects the RMSE value. In Fig. 10, the RMSE of different geographical regions at different peak times is shown. The area of this geographical area is 0.5, 1.0, 1.5 square kilometers respectively. It is easy to understand that as the area increases, the RMSE value increases. Because when the area of each block is smaller, the number of blocks will be more, and the higher clarity of the historical snapshot will be received. Therefore, the snapshot contains more information. In order to improve the accuracy of the framework, the smaller area partitioning method should be chosen.

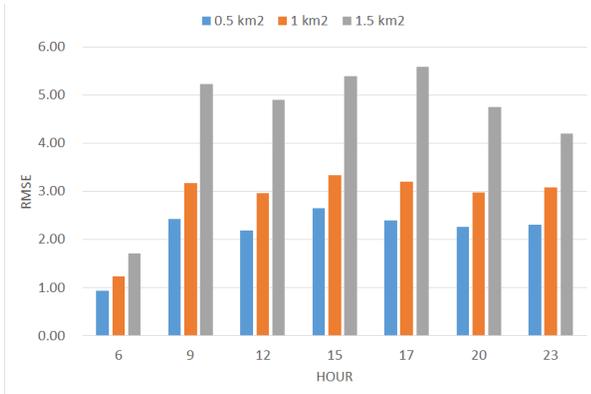

Fig. 10. RMSE in different sizes of the partition at different rush hours.

But if the area is too small, the time complexity of the traffic prediction framework will be high. In Fig. 11, we compare the efficiency of the model with different size blocks at different rush times. It finds that if the blocks are smaller, the model has a longer execution time. At the same time, if the partitioning method is the same, the time performance of the model has little change. Because the block area is smaller, the number of the block will be larger. When doing convolution operation, it will waste much time. So we should consider the time cost and forecast accuracy simultaneously to choose an appropriate partitioning method for travel demand prediction.

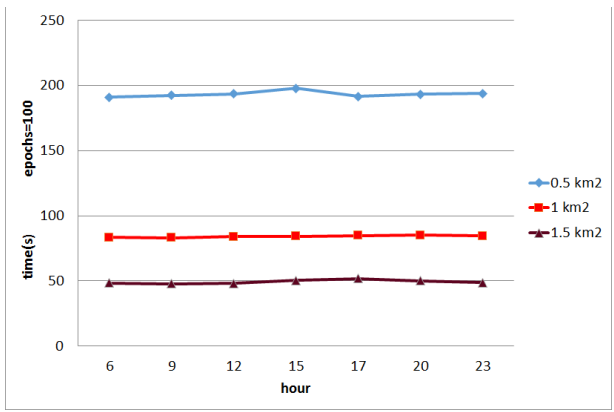

Fig. 11. The time performance in different size of the zone (epoch = 100)

## V. CONCLUSION

Travel demand modeling is an inherent part of smarter transportation. Analyzing and forecasting travel demand can help us manage the hot spot of passenger demand in the next period, balance supply and demand and schedule vehicle resources for passengers.

In this paper, a ConvLSTM-based deep learning model for travel demand (ST Data) prediction is proposed that takes advantage of both temporal and spatial properties. We evaluate our models on a real-world dataset (DIDI GAIA Order dataset). Our models' performances are significantly beyond two baseline models, confirming that it is better and more flexible for the travel demand prediction. In addition, we found that, a smaller block brings accurate result but wastes more time. So we should choose a suitable partitioning method to predict travel demand.

In the future, we will consider adding incremental learning into our deep learning model. When we get real-time data, our model can adapt to different data and adjust its own parameters by itself.